# Learning Based on CC1 and CC4 Neural Networks

Subhash Kak
Oklahoma State University

*Abstract*— We propose that a general learning system should have three kinds of agents corresponding to sensory, short-term, and long-term memory that implicitly will facilitate context-free and context-sensitive aspects of learning. These three agents perform mututally complementary functions that capture aspects of the human cognition system. We investigate the use of CC1 and CC4 networks for use as models of short-term and sensory memory.

I. INTRODUCTION

The human cognition system works at two levels: it analyzes surface structure of an utterance using grammatical rules but makes deductions only within the framework of the underlying context. Similarly, implicit learning and memory of visual context can guide spatial attention towards task-relevant aspects of the scene [1].

The context-independent properties are not affected by context, whereas context-dependent properties are. Context-independent properties constitute the core meaning, whereas context-dependent properties lead to semantic encoding variability [2]. In discourse analysis one needs to know the context before one can analyze a specific event [3].

Patterns that might appear distinct at one level of analysis may not look so when analyzed at another level. This idea should be useful in the problem of neural network training. We propose this can be done most easily using corner classification neural networks (CCNN) for dynamic classification of data sets [4][5][6][7] and for modeling of memory.

Memory may be divided into three types, sensory, short-term and long-term memory. The duration for which information can be retained is shortest for sensory memory and longest for greatest for long-term memory, and short-term memory stands in between the sensory and long-term memory.

Short-time memory, also called working memory, fades approximately after twenty seconds if it is not renewed through rehearsal. It needs to be protected from overloading by sensory stimulation, two cognitive processes that help in preventing overloading are sensory gating and selective attention. Sensory gating is the process by which certain channels are turned on while others are turned off. The amount of information that short-term memory can hold is limited but it can be extended by "grouping" information. Both of these must use very fast high-level classification that operates on sensory memory.

It was proposed that CCNN networks can serve as models for short-term memory [8], however no attempt was made to distinguish between sensory and the more formally called short-term memory, and this distinction will be made for the first time in the present paper.

In the CCNN, the basis idea is to take the training sample and associate a δ-neighborhood (consisting of all points within the radius of δ from the training sample) to the same class The training can adapt quickly to the changes in the underlying data stream. But it leaves out the question of the choice of δ to the designer and the nature of the application.

CCNN trained neural networks may be used in two basic modes: (i) where the δ-value adapts to the learning sample (CC1), and (ii) where it is fixed (CC4). Clearly, There is a burden of computation required in CC1 (for the adaptation to come through and this may require many rounds of adjustment), whereas little computation is required in CC4. For the advantage of computation requirements, CC4 networks have become popular and they have been used in many applications ranging from document classification to time-series prediction [9][10][11][12].

Either CC1 or CC4 could be used as a model of short-term ephemeral memory. One might assume that this memory is based originally on CC4 and further refinement is performed by doing a CC1 run on it to reduce error. Such a memory can be used in tandem with deep learning models [13][14][15], where the specific deep learning network is determined by the application at hand.

There may be other aspects of learning that are non-classical [16][17][18], which we shall not go into in this paper. These elements might help to improve the performance of deep neural networks which suffer from certain pathologies (such as classifying unrecognizable images as belonging to a familiar category of ordinary images and misclassifying minuscule perturbations of correctly classified images [19],[20],[21]).

This paper further investigates general properties of corner classification architecture using standard learning. We propose that CC4 and CC1 networks can represent sensory and short-term memory, a distinction that is being made in relation to artificial neural networks for the first time. This paper does not go into the exact manner in which sensory and short-term memory may be used in an actual cognitive task and that is a problem that needs to be further examined. Here we only say that we expect that both sensory and short-term memory ideas when used in juxtaposition with newer algorithms of deep learning will offer improved performance.

## II. CORNER CLASSIFICATION LEARNING

The *corner classification* (CC) network is based on the idea of phonological loop and the visio-spatial sketchpad [8] from the perspective of neuroscience, and separating corners of a hypercube from the perspective of machine learning [4][5]. There are four versions of the CC technique, represented by CC1 through CC4 but only these two are of significance now since CC2 and CC3 were intermediate stages that led to the development of CC4.

These networks are a feedforward network architecture consisting of three layers of neurons. The number of input neurons is equal to the length of input patterns or vectors plus one, the additional neuron being the bias neuron, which has a constant input of 1. The number of hidden neurons is equal to the number of training samples, each hidden neuron correspond to one training example. In more advanced versions, the hidden neurons may eventually be pruned down.

Each node in the network acts as a filter for the training sample. The filter is realized by making it act as a hyperplane to separate the corner of the *n*-dimensional cube represented by the training vector and hence the name corner-classification technique.

The CC1 network optimizes the generalization for each learning sample and is, therefore, an adaptive network that may also be seen from the perspective of a network where the radius of generalization varies with the training sample.

In the CC4, the last node of the input layer is set to one to act as a bias to the hidden layer. The binary step function is used as the activation function for both the hidden and output neurons. The output of the activation function is 1 if summation is positive and zero otherwise.

For each training vector presented to the network, if an input neuron receives a 1, its weight to the hidden neuron corresponding to this training vector is set to I. Otherwise, it is set to -1. The bias neuron is treated differently. If $s$ is the number of 1's in the training vector, excluding the bias input, and the desired radius of generalization is $r$, then the weight between the bias neuron and the hidden neuron corresponding to this training vector is $r - s + 1$.

The weights in the output layer are equal to 1 if the output value is 1 and –1 if the output value is 0. This amounts to learning both the input class and its complement and thus instantaneous. The radius of generalization, $r$ can be seen by considering the all-zero input vectors for which $w_{n+1} = r + 1$.

The choice of $r$ will depend on the nature of generalization sought. Since the weights are 1, -1, or 0, it is clear that actual computations are minimal. In the general case, the only weight that can be greater in magnitude than 1 is the one associated with the bias neuron. When real data is represented in binary, that should be done using *unary* coding which is the coding biological learning also appears to be based on [22][23][24].

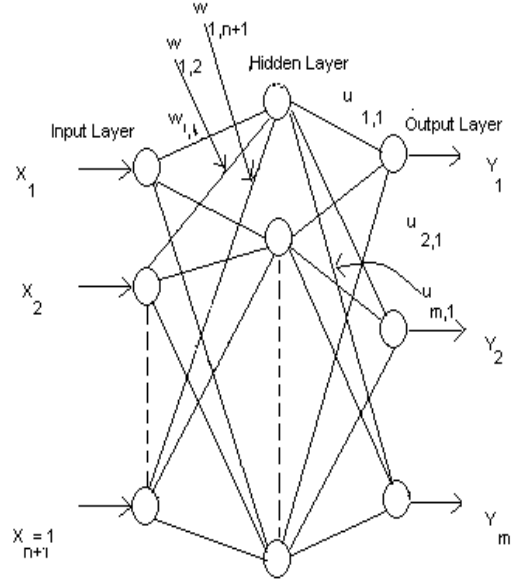

**Figure 1**: General CC architecture

The varying radius generalization architecture of CC1 may be represented by Figure 2.

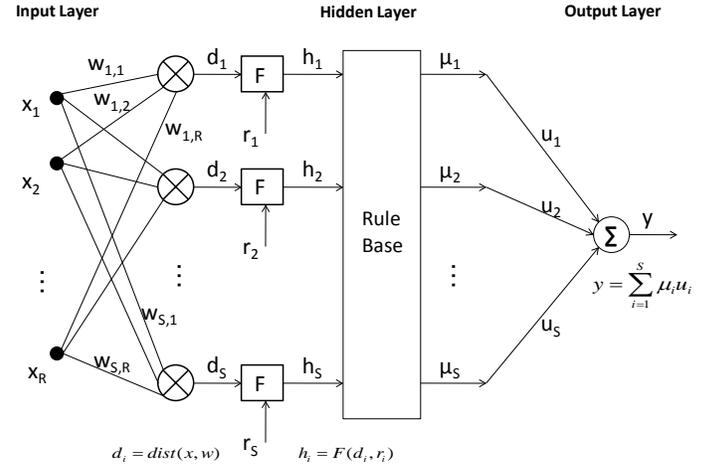

**Figure 2.** CC1 realization

The CC1 network of Figure 2 compares the stored vectors of the data to the input data using an appropriate distance metric. The input data is normalized and presented as input vector $x$. The hidden neuron is represented by the weight vector $w_i$ and its elements are represented by $w_{i,j}$, $i = (1,2,…,S)$ and $j = (1,2,…,R)$, where $R$ is the number of components of the input vector and $S$ is the number of hidden neurons (the number of training samples). The output is the dot product of the vectors $\mu$ and $u$, where $\mu$ is the vector at the output of the Rule Base and $u$ is the vector of weights in the output layer as shown in Figure 2. This network can be trained with just two passes of the samples, the first pass assigns the synaptic weights and the second pass determines the radius of generalization for each training sample [7].

The network behaves as a 1NN classifier and a kNN classifier (1- or k- nearest neighbor classifier) according to whether the input vector falls within the radius of generalization of a training vector or not. The radius of generalization acts as a switch between the 1NN classifier and the kNN classifier.

This network meets the specifications set by traditional function approximation that every data point is covered in the given training sample. In the practical case, the *k* values are determined by the sample size and be a fraction of the sample size. If k=S then the FC network operating as a kNN classifier can be viewed as a RBF network provided the membership function is chosen to be a Gaussian distributed. On the other hand, if the weighting function is chosen to be the membership function, the CC1 classifier can be considered as kernel regression. As in the CC networks, this network requires as many hidden neurons as the number of training samples (although the number of hidden neurons could be trimmed to a certain extent).

### III. ERROR CHANGE WITH TRAINING SET SIZE

We now compare the error performance of CC1 and CC4 networks. We do this in the context of a pattern classification problem like that of Figure 3, where the shapes belong to a single class which is distinct from the background. The problem is to learn this class and generalize in the process by using a fraction of the elements of the picture.

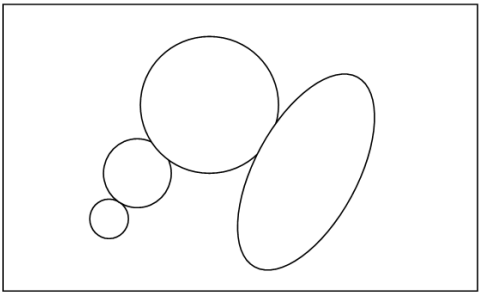

**Figure 3.** Example classification problem

In the CC1 network, one requires a comparison with stored information in one pass, and, therefore, it is not quite instantaneous, but the calculation can be done in time that could be smaller than the time instant at which the next data comes in. The CC4 network training is done using different values of the radius of generalization.

Figure 4 presents a comparison between the error for the CC1 and CC4 training. Since the CC1 network adapts the generalization region based on the distances between the training samples, the reduction in error as the number of training samples increases is not very dramatic for the structured shapes of Figure 3.

On the other hand, the error values go down as the proportion of training samples increases. The CC4 training for each number of the X-axis is assumed to use the best possible value of radius of generalization for that set of the training samples. As the size of the training set becomes larger the error using CC4 comes down closer to the CC1 error.

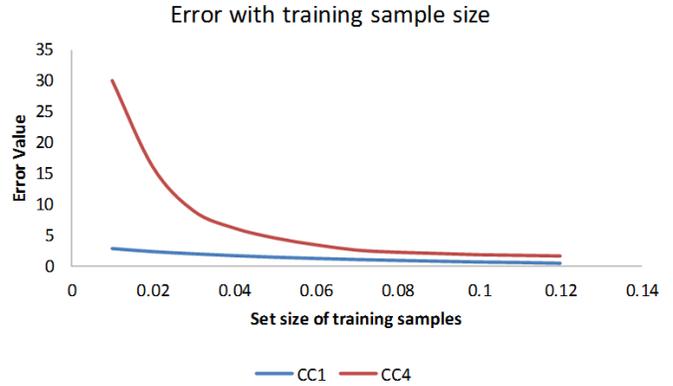

**Figure 4.** Comparative error performance of CC1 and CC4

The error in CC4 is a minimum for some intermediate value of r. When it is very small, the generalization is inadequate and there are numerous gaps in the learning. On the other hand, if r is made too large, the generalization is not precise enough to separate regions that are near, which leads to accumulation of error.

This relationship of error in generalization with respect to r for a CC4 network is shown in Figure 5.

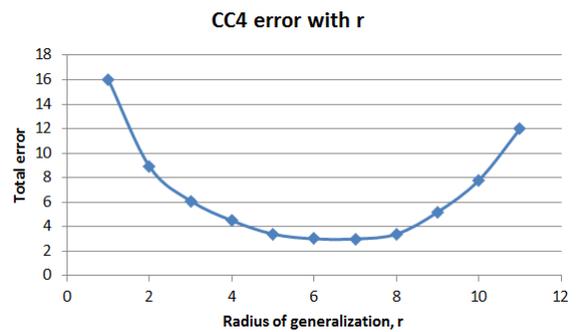

**Figure 5.** CC4 error rate with radius of generalization

### IV. NUMBER OF CLASSES

The number of classes that CC networks can distinguish between depends on the number of neurons in the output layer.

The neurons in the output layer serve as decision agents in each stage of a decision tree. For example, Figure 6, which defines the working of the output layer for 3 neurons is able to separate 8 regions. Effectively, each neuron makes a binary decision and taken together, one is able to account for $2^k$ classes if one has k output layer neurons.

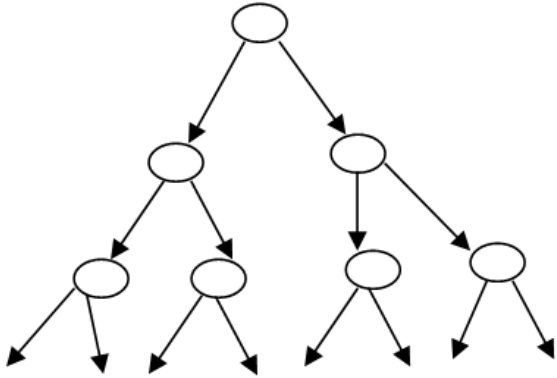

**Figure 6.** Decision layers in the output neurons

## V. TIME-SERIES PREDICTION

In the use of CC4 network for prediction, the choice of radius of generalization would depend on the nature of the data and the size of the training set. Basically, prediction is done using a moving window. The network is trained by the data that is already at hand and it is then used to predict the future value. One can see how such a method will be useful in financial time-series prediction where good prediction can be converted into financial advantage.

Mackey-Glass (MG) time series which is based on Mackey-Glass differential equation is often used as a test sequence for checking how good the prediction technique is.

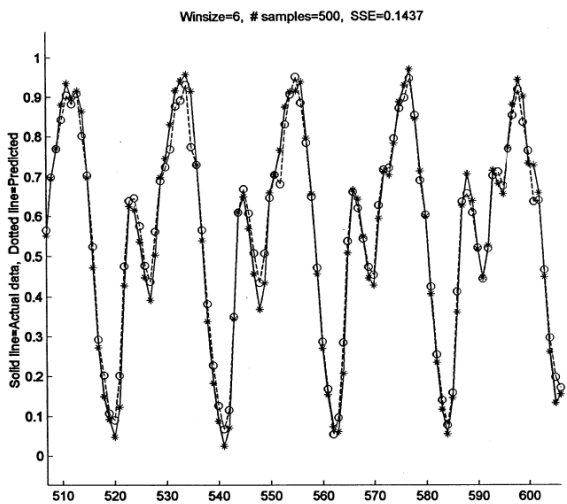

**Figure 7.** Prediction of chaotic series using CC1

The discrete time representation of the Mackey-Glass (MG) equation is:

$$x(t + 1) = (1 − B)x(t) + Ax(t − D)/[1 + x^C(t − D)]$$

where A, B, and C are constants, and D is the time delay parameter. Under a suitable choice of these numbers, the resultant time series will exhibit chaotic behavior. The popular case with A = 0.2, B = 0.1, C = 10, and the delay parameter D set to 30 is selected here.

In time-series prediction another significant parameter is the number of past values that will be used for determining the next value and we call it k. In other words, the window size is k and prediction is made for one point ahead. The CC1 prediction method was called FC prediction in [7].

The MG equation was used to generate a continuous sequence of data points. The first 3,000 points were discarded to allow initialization transients to decay. The remaining data points were sampled once every six points to obtain the actual time series used for this experiment. A total of 500 samples are used for training. It was found that the error was very small and nearly the same for k = 5 and 6. Figure 7 gives a plot of the time series and its prediction and one can see how close the one-point ahead prediction is to the actual waveform.

## VI. CONCLUSION

This paper proposed that a general system should have three kinds of learning agents corresponding to sensory, short-term, and long-term memory. These three agents perform complementary functions that can improve the overall performance in a manner similar to human cognitive agent. This paper investigated the use of CC4 and CC1 networks for use as models of short-term and sensory memory.